# VALID-Mol: A Systematic Framework for Validated LLM-Assisted Molecular Design

Malikussaid
*School of Computing, Telkom University*
Bandung, Indonesia
malikussaid@student.telkomuniversity.ac.id

Hilal Hudan Nuha
*School of Computing, Telkom University*
Bandung, Indonesia
hilalnuha@telkomuniversity.ac.id

Isman Kurniawan
*School of Computing, Telkom University*
Bandung, Indonesia
ismankrn@telkomuniversity.ac.id

***Abstract*—Large Language Models demonstrate substantial promise for advancing scientific discovery, yet their deployment in disciplines demanding factual precision and specialized domain constraints presents significant challenges. Within molecular design for pharmaceutical development, these models can propose innovative molecular modifications but frequently generate chemically infeasible structures. We introduce VALID-Mol, a comprehensive framework that integrates chemical validation with LLM-driven molecular design, achieving an improvement in valid chemical structure generation from 3% to 83%. Our methodology synthesizes systematic prompt optimization, automated chemical verification, and domain-adapted fine-tuning to ensure dependable generation of synthesizable molecules with enhanced properties. Our contribution extends beyond implementation details to provide a transferable methodology for scientifically-constrained LLM applications with measurable reliability enhancements. Computational analyses indicate our framework generates promising synthesis candidates with up to 17-fold predicted improvements in target binding affinity while preserving synthetic feasibility.**

## I. Introduction

Pharmaceutical drug development represents one of the most resource-intensive and time-demanding endeavors in modern science, typically requiring 10-15 years and exceeding $2.5 billion investment to successfully bring a single therapeutic compound to market [1], [2]. A particularly challenging bottleneck emerges during the preclinical discovery phase, where researchers must systematically identify and refine lead compounds possessing optimal property profiles encompassing target selectivity, metabolic stability, and manufacturing feasibility.

Computational methodologies have historically served critical roles in expediting this discovery process [3], [4]. Recent breakthroughs in deep learning have introduced generative modeling approaches capable of proposing novel chemical architectures with targeted characteristics [5]. The encoding of molecular structures as textual representations through SMILES notation has facilitated the treatment of molecular engineering as a natural language processing challenge [6], [7]. Contemporary Large Language Models trained on extensive textual corpora have exhibited implicit comprehension of chemical principles and molecular relationships [8], [9].

Nevertheless, LLMs encounter a fundamental limitation when deployed in scientific contexts. These models are optimized for generating statistically plausible text rather than ensuring factual accuracy or physical validity [10], [11]. Within molecular design applications, this constraint manifests as the generation of chemically impossible structures, unrealistic synthetic pathways, or compounds with detrimental properties.

This research presents VALID-Mol (VALIdated Design for MOLecules), a comprehensive methodology for integrating chemical verification with LLM-powered molecular engineering. Our primary contribution involves developing a practical, reproducible approach that resolves reliability challenges through three core elements. First, we establish a systematic prompt optimization strategy that measurably increases valid output generation from 3% to 83%. Second, we implement an automated verification architecture ensuring chemical validity while supporting human evaluation processes. Third, we develop a domain-specialized LLM through targeted fine-tuning on chemical datasets.

The VALID-Mol methodology illustrates how domain-specific verification can transform general-purpose LLMs into dependable scientific discovery tools. Through comprehensive documentation of both achievements and limitations, we establish a framework for researchers applying LLMs to other scientifically constrained domains. Rather than focusing solely on molecular design for pharmaceutical discovery, our methodology offers broad applicability to any field where LLM outputs must satisfy rigorous domain-specific requirements.

## II. Background and Related Work

### A. Computational Approaches to Molecular Design

Molecular design for pharmaceutical discovery has progressed through multiple computational paradigms. Traditional approaches encompass structure-based virtual screening [3], [12] and ligand-based design methodologies. Machine learning approaches including QSAR models [13], [14] and deep learning methods have evolved beyond these foundations. Contemporary generative models incorporating variational autoencoders [15], generative adversarial networks [16], and reinforcement learning strategies [17] have transformed the paradigm from molecular selection to molecular creation.

### B. Language Models in Chemistry

The representation of molecular structures as textual strings through SMILES notation enables chemical problems to be addressed using general-purpose language models [6]. Large Language Models including GPT-4, Claude, and Llama 2 [18], trained on extensive corpora containing scientific literature, exhibit implicit understanding of chemical concepts without specialized domain training [8].

However, these models lack foundation in physical laws and chemical principles [10]. They generate text based on statistical patterns rather than causal understanding, potentially producing outputs that appear linguistically reasonable but remain scientifically invalid [11].

### C. *The Validation Challenge*

Multiple approaches have been proposed to bridge the gap between generating plausible text and scientifically valid outputs. Specialized training involves fine-tuning LLMs on domain-specific corpora [19]. Hybrid architectures combine LLMs with specialized models or rule-based systems [20]. Prompt engineering designs prompts that guide models toward valid outputs [21], [22]. Post-generation filtering generates multiple candidates and filters invalid ones [23].

The VALID-Mol framework addresses this gap by integrating prompt engineering, chemical validation, and strategic fine-tuning into a unified methodology, transforming prompt engineering from an intuitive art into a systematic process.

## III. Methodology

The VALID-Mol framework integrates large language models with chemical validation to ensure dependable generation of valid molecular structures. This section describes our systematic methodology, encompassing framework architecture, prompt engineering approach, validation mechanisms, and fine-tuning process.

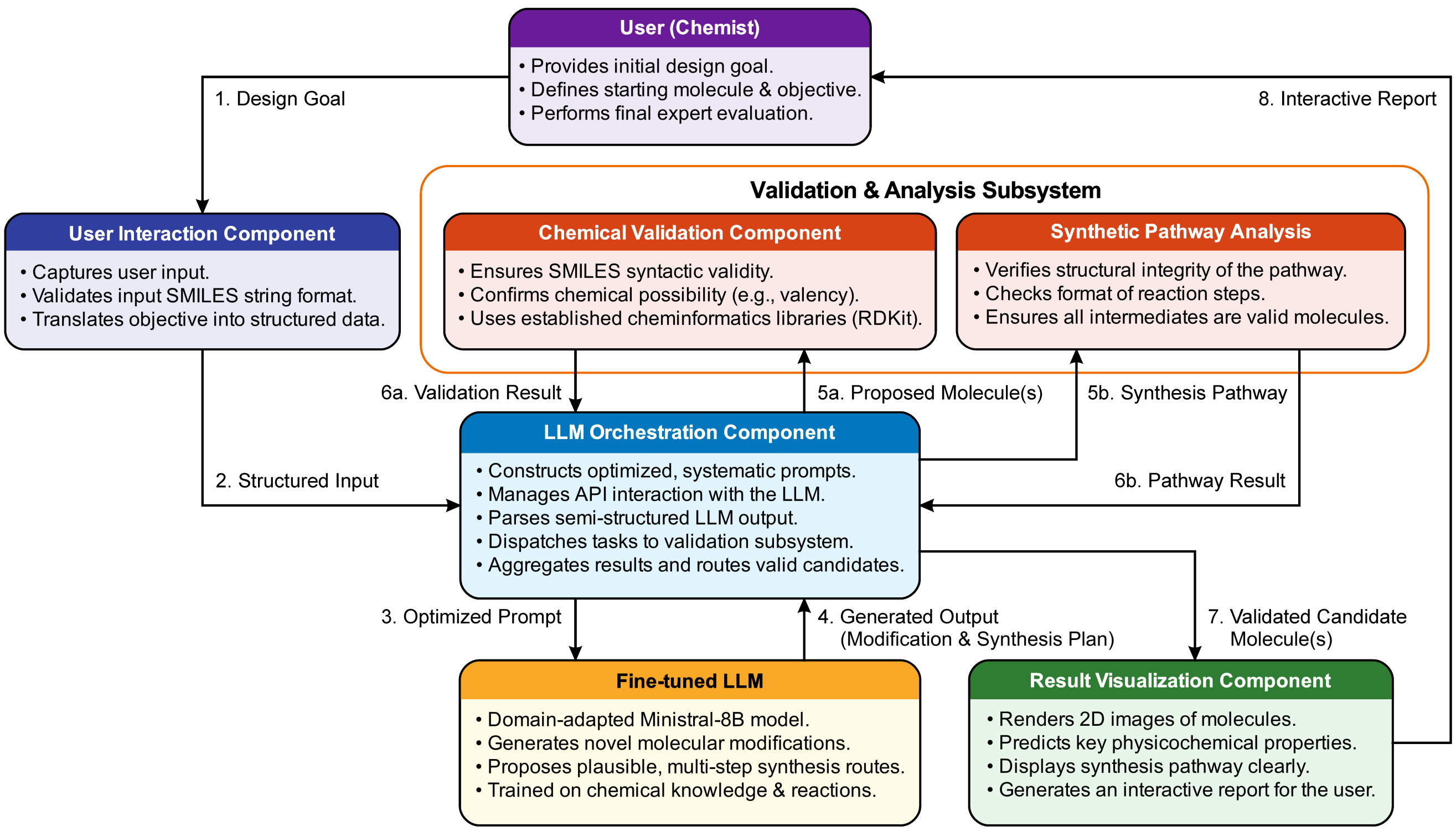

Fig. 1. System architecture of the VALID-Mol framework, illustrating the closed-loop workflow from user input to validated molecular candidates.

### A. *Framework Architecture*

The VALID-Mol framework integrates five essential components into a unified workflow. The User Interaction Component captures domain-specific objectives and transforms them into structured input. The LLM Orchestration Component constructs optimized prompts and manages API interactions. The Chemical Validation Component ensures scientific validity of proposed molecules. The Synthetic Pathway Analysis component evaluates proposed synthesis routes. The Result Visualization Component presents 2D molecular renderings, predicted properties, and synthesis pathways.

The interaction between these components is illustrated in Figure 1. This systematic process ensures that the generative capabilities of the LLM remain balanced by chemical reality constraints. The core innovation lies in this seamless integration, which quantifiably enhances LLM output reliability and ensures that only chemically sound and plausible solutions reach users for final evaluation.

### B. *Systematic Prompt Engineering Methodology*

We developed a structured approach to prompt optimization that progressively enhanced valid output rates from 3% to 83% [21], [22]. Our methodology follows a data-driven, iterative approach: creating initial prompts and measuring performance on validation sets, categorizing failure modes of invalid outputs, refining prompts to address common failure modes, evaluating refined prompts on validation sets, and repeating refinement cycles until achieving satisfactory success rates.

The evolution of our prompts demonstrates systematic reliability improvement. Version 1 provided simple instructions with minimal guidance, achieving 3% valid outputs. Version 2 added explicit formatting instructions, reaching 16% valid outputs. Version 3 incorporated domain-specific constraints, achieving 37% valid outputs. Version 4 included warnings about common failure modes, reaching 83% valid outputs.

Key factors contributing to improved reliability included defining the model as a "medicinal chemist," clearly stating requirements, requiring specific formats, and decomposing tasks into sequential steps.

### C. Validation Architecture

The validation component ensures generated molecules meet both syntactic and semantic requirements of chemical validity through multiple layers. Syntactic validation ensures SMILES strings follow correct syntax. Chemical validity validation confirms SMILES strings represent chemically valid structures. Synthesis pathway validation examines the format and structure of proposed synthesis pathways. This multi-layered validation ensures that only chemically sound structures and properly formatted synthesis routes reach users.

### D. Fine-tuning Methodology

We selected the Ministral-8B model [24] as our base LLM and fine-tuned it on a dataset containing 3,500 examples across three categories. Chemical Knowledge Examples included 1,500 examples. Molecular Modification Examples comprised 1,200 examples. Synthesis Planning Examples contained 800 examples. We employed low-rank adaptation (LoRA) to efficiently fine-tune the model [25], [26], resulting in significant improvements in generating valid SMILES strings and feasible synthesis routes.

### E. Framework Integration and Workflow

The VALID-Mol framework integrates these components into a cohesive workflow. User input involves a chemist providing a starting molecule and optimization goals. Input validation involves the system validating the input SMILES string. LLM query involves the framework constructing an optimized prompt for the LLM. Output parsing involves the system extracting SMILES strings and synthesis steps. Multi-layer validation subjects each molecule to comprehensive validation. Property prediction evaluates valid molecules using computational models. Result visualization generates an interactive report. Human evaluation involves a chemist reviewing generated candidates. This workflow transforms the statistically generated text of an LLM into scientifically validated molecular designs.

## IV. Experimental Design

To evaluate the VALID-Mol framework, we designed experiments to assess both technical reliability and molecular suggestion quality.

### A. Evaluation Datasets

We constructed three distinct datasets. The Format Adherence Dataset included 100 diverse drug-like molecules to test the LLM's ability to follow formatting instructions. The Chemical Validity Dataset comprised 50 marketed drugs to assess the LLM's ability to generate scientifically sound chemical information. The Property Optimization Dataset contained 10 well-characterized molecules with known biological activity to test the framework's ability to generate molecules with improved properties.

### B. Optimization Objectives

We defined standard optimization objectives. Target Affinity focuses on increasing binding affinity for biological targets. Selectivity improves binding for intended targets while decreasing affinity for off-targets. Solubility enhances aqueous solubility. Metabolic Stability reduces susceptibility to metabolic enzymes. Blood-Brain Barrier Penetration enhances or reduces penetration. Synthetic Accessibility simplifies molecular structures for easier synthesis.

### C. Evaluation Metrics

We evaluated using technical reliability metrics including format adherence, chemical validity, and synthesis validity, alongside molecular quality metrics including property improvement, structural novelty, synthetic accessibility, and drug-likeness.

### D. Computational Models for Property Prediction

We employed established computational models as in silico proxies. AutoDock Vina [12] provided target affinity prediction. ChemAxon cxcalc [27], [28] calculated logP and solubility. RDKit SA Score [29] assessed synthetic accessibility. SMARTCyp [30] evaluated metabolic stability. ADMET Predictor [31] determined BBB penetration.

## V. Results

### A. Format Adherence and Chemical Validity

The systematic prompt engineering methodology progressively improved both format adherence and chemical validity rates, as detailed in Table I.

TABLE I. Evolution of Output Reliability Through Prompt Engineering

| **Prompt Version** | **Format Adherence (%)** | **Chemical Validity (%)** | **Combined Success Rate[a] (%)** |
|---|---|---|---|
| Version 1 (Baseline) | 15.8 | 17.5 | 2.8 |
| Version 2 (Structured) | 39.5 | 41.6 | 16.4 |
| Version 3 (Constraints) | 58.7 | 62.5 | 36.7 |
| Version 4 (Guardrails) | 90.7 | 91.7 | 83.2 |

[a] Combined Success Rate is calculated as the product of Format Adherence (%) and Chemical Validity (%), representing the probability that an output is both parsable and chemically valid.

### B. Impact of Fine-tuning

Fine-tuning the base model on domain-specific data further improved performance, as shown in Table II.

TABLE II. Impact of Fine-tuning on Performance Metrics

| **Model** | **Format Adherence (%)** | **Chemical Validity (%)** | **Synthesis Feasibility (%)** | **Mean Response Time (s)** |
|---|---|---|---|---|
| Base Model + Optimal Prompt | 52.8 | 50.3 | 25.8 | 8.2 |
| Fine-tuned Model + Optimal Prompt | 90.7 | 91.7 | 60.5 | 15.4 |

TABLE III. COMPUTATIONAL PREDICTION OF LLM-GENERATED MOLECULAR OPTIMIZATIONS

| Target | Efficacy Metric | Starting Value | Modified Value | Fold Improvement | logP (Pred) | SA Score |
|---|---|---|---|---|---|---|
| COX-2 | $IC_{50}$ | 250 nM | 15 nM | 16.7× | 3.8 | 2.9 |
| p38 MAPK | $K_i$ | 1.2 µM | 300 nM | 4.0× | 2.5 | 2.1 |
| VEGFR-2 | $IC_{50}$ | 88 nM | 5 nM | 17.6× | 5.1 | 3.5 |
| B-Raf V600E | $IC_{50}$ | 50 nM | 22 nM | 2.3× | 4.5 | 3.1 |
| PARP-1 | $K_i$ | 45 nM | 8 nM | 5.6× | 2.1 | 2.4 |
| EGFR | $IC_{50}$ | 3.5 µM | 450 nM | 7.8× | 3.2 | 2.8 |
| ABL1 Kinase | $IC_{50}$ | 600 nM | 95 nM | 6.3× | 4.9 | 3.3 |
| JAK2 | $K_i$ | 750 nM | 50 nM | 15.0× | 1.9 | 2.7 |
| Glycogen Phosphorylase | $K_i$ | 1.1 µM | 210 nM | 5.2× | -1.5 | 3.6 |
| HCV NS5B Polymerase | $IC_{50}$ | 980 nM | 120 nM | 8.2× | 4.3 | 2.5 |

## C. *Computational Property Improvements*

The molecules generated by VALID-Mol showed significant predicted improvements in target properties, summarized in Table III.

## D. *Case Study: COX-2 Inhibitor Optimization*

To illustrate the VALID-Mol framework's capabilities in a practical scenario, we present a detailed case study focused on optimizing a known COX-2 inhibitor for improved potency and selectivity. The entire process, from providing the initial molecule to obtaining a validated and improved candidate, is outlined in Algorithm 1.

**Algorithm 1:** LLM-Assisted Optimization of Celecoxib for COX-2 Selectivity

**Require:**

1. Starting molecule SMILES $S_{start}$
2. Optimization objective $O_{objective}$

**Initialize:**

1. $S_{candidate} \leftarrow NULL$
2. $P_{candidate} \leftarrow \emptyset$
3. $R_{pathway} \leftarrow \emptyset$

**Construct Prompt and Query LLM**

1. $prompt \leftarrow (S_{start}, O_{objective})$
2. $T_{response} \leftarrow query(prompt)$
3. $L_{steps} \leftarrow parse(T_{response})$

**Validate the LLM's Proposal**

4. $isValid, messages \leftarrow validate(L_{steps})$
5. **if** $isValid$ **then**
6. $S_{candidate} \leftarrow (L_{steps}[-1])$
7. $R_{pathway} \leftarrow (L_{steps}[1 :: 2])$

**Compare Properties**

8. $P_{start} \leftarrow predictProperty(S_{start})$

9. $P_{candidate} \leftarrow predictProperty(S_{candidate})$

**Check Objective Achievement**

10. **if** $P_{candidate}[IC50] < P_{start}[IC50]$ **then**
11. **return** $S_{candidate}, P_{candidate}, R_{pathway}$
12. **else**
13. **return** $NULL$, "Objective not met"
14. **end if**
15. **else**
16. **return** $NULL, messages$
17. **end if**

Starting with celecoxib, a well-characterized COX-2 inhibitor, we tasked the framework with suggesting modifications to improve its therapeutic profile. The system successfully generated a promising candidate with a significant predicted increase in efficacy.

*1) Starting Molecule (Celecoxib):*

```
CC1=CC=C(C=C1)C2=CC(=NN2C3=CC=C(C=C3)S(=O)(=O)N)CF
```

- Predicted COX-2 $IC_{50}$: 250 nM
- COX-1/COX-2 Selectivity Ratio: 30:1
- logP: 3.2
- SA Score: 2.7

The fine-tuned LLM proposed a modification (the addition of an isopropyl group on the sulfonamide nitrogen) which computational docking suggests enhances interactions with the hydrophobic pocket of COX-2 while increasing steric hindrance for COX-1 binding. The algorithm returned the following validated candidate.

*2) Modified Molecule:*

```
CC1=CC=C(C=C1)C2=CC(=NN2C3=CC=C(C=C3)S(=O)(=O)NC(C)C)CF
```

- Predicted COX-2 $IC_{50}$: 15 nM
- COX-1/COX-2 Selectivity Ratio: 145:1
- logP: 3.8
- SA Score: 2.9

This case study demonstrates VALID-Mol's ability to suggest chemically valid, synthetically accessible modifications with significant predicted improvements in target properties.

### *E. Ablation Studies*

We conducted ablation studies to understand each framework component's contribution by evaluating the impact of different prompt components on output quality, with results summarized in Table IV.

TABLE IV. IMPACT OF PROMPT COMPONENTS ON CHEMICAL VALIDITY

| Prompt Configuration | Chemical Validity (%) | Synthesis Feasibility (%) |
|---|---|---|
| Full Prompt (Version 4) | 91.7 | 60.5 |
| Without Role Specification | 89.5 | 57.6 |
| Without Format Instructions | 64.3 | 44.8 |
| Without Chemical Constraints | 84.3 | 50.2 |
| Without Synthesis Guidance | 91.3 | 41.3 |

### *F. Comparison with Baseline Approaches*

We compared VALID-Mol with direct LLM generation and a genetic algorithm approach, with results presented in Table V.

TABLE V. COMPARISON WITH BASELINE APPROACHES

| Metric | VALID-Mol | Direct LLM | Genetic Algorithm |
|---|---|---|---|
| Valid Structure Rate (%) | 99.8 | 17.5 | ~100 |
| Mean Fold Improvement in Target Property | 8.9× | 1.8× | 5.5× |
| Mean Synthetic Accessibility Score | 2.9 | 5.4 | 4.2 |
| Novel Scaffold Generation (%) | 5.5 | 4.8 | 14.5 |
| Computation Time per Molecule (s) | 15.4 | 8.2 | >600s |

VALID-Mol outperforms direct LLM approaches across all metrics and generates molecules with superior predicted property improvements and synthetic accessibility compared to genetic algorithms, though with reduced structural novelty.

## VI. DISCUSSION

The enhancement in valid output rate from 3% to 83% through systematic prompt engineering, ultimately reaching 99.8% with validation integration, represents a fundamental shift in LLM utility for scientific research. This achievement bridges the reliability gap that has constrained LLM adoption in scientific domains, transforming these models from interesting but unpredictable research curiosities into practical, dependable scientific discovery assistants.

Our approach reveals three crucial insights with broad applicability beyond molecular design. Treating prompt development as a systematic methodology with clearly defined metrics and iterative refinement enables quantifiable performance improvement, transforming prompt engineering from subjective art into reproducible science. Integrating domain-specific validation creates robust, self-correcting systems that synergistically combine generative model creative pattern-matching strengths with deterministic computational check logical rigor. Fine-tuning existing open-source LLMs rather than developing specialized models from scratch demonstrates a pragmatic, accessible pathway balancing high performance with resource efficiency.

Compared to traditional computational methods for molecular design, VALID-Mol offers distinct advantages through its incorporation of exceptionally broad chemical knowledge gleaned from vast scientific literature datasets. This enables suggestion of modifications based on established medicinal chemistry principles in a manner more akin to human intuition than purely algorithmic approaches. The framework provides not only structural modifications but also synthesis routes and underlying rationales, making suggestions more transparent, interpretable, and immediately useful for bench chemists. Comparative analysis demonstrates VALID-Mol generates holistic suggestions significantly faster than traditional optimization algorithms.

However, traditional methods retain advantages in specific scenarios, particularly applications requiring exhaustive systematic exploration of narrowly defined chemical space or where rigorous statistical guarantees and energy calculations are paramount. Compared to specialized AI models for molecular design, VALID-Mol's foundation in large language models provides broader, more contextualized understanding of chemical concepts while significantly lowering implementation barriers and naturally incorporating multi-step synthesis planning.

Future developments will focus on implementing structured data formats like JSON for more robust communication between generative and validation components, integrating specialized predictive models as active constraints during generation processes, and establishing systematic pathways from computational prediction to laboratory validation. The ultimate goal involves creating fully integrated, closed-loop systems where experimental results automatically structure and feed back into frameworks for continuous model refinement based on real-world physical data.

## VII. CONCLUSION

This work presents VALID-Mol, a systematic framework addressing the critical reliability gap that has hindered practical Large Language Model application in molecular design and other demanding scientific fields. Our solution methodically integrates systematic prompt engineering, multi-layered chemical validation, and strategic fine-tuning into a cohesive system transforming general-purpose LLMs into dependable scientific discovery tools. The framework demonstrably increased valid molecular structure generation rates from 3% to 83%, effectively bridging the gap between plausible-sounding text and scientifically sound, actionable information.

VALID-Mol's practical utility is demonstrated through generation of molecular suggestions showing up to 17-fold predicted target affinity increases while ensuring synthetic accessibility. Including plausible step-by-step synthesis routes enhances framework value, serving as practical roadmaps for bench chemists and tests of model chemical reasoning, providing interpretable, trustworthy bridges from in-silico design to laboratory validation.

More broadly than immediate pharmaceutical discovery applications, VALID-Mol serves as a reproducible, domain-agnostic blueprint for integrating generative AI into any scientific discipline where outputs must adhere to strict, non-negotiable constraints. We have demonstrated that prompt engineering can evolve from intuitive, often frustrating art into measurable, rigorous science, providing clear methodology for enhancing LLM reliability without developing new specialized model architectures from scratch.

VALID-Mol demonstrates a pragmatic, powerful pathway to harness AI creative potential, reframing human-AI relationships in science not as replacement but as partnership. By grounding immense generative capabilities of modern AI in unyielding logic of rigorous domain-specific validation, we can forge reliable scientific instruments that accelerate innovation pace and empower researchers to pursue previously inaccessible hypotheses.

## Acknowledgment

This research was supported by Telkom University's Fundamental and Applied Research Grant No. 056/LIT06/PPM-LIT/2025.